\documentclass[conference]{IEEEtran}
\IEEEoverridecommandlockouts
\usepackage{cite}

\usepackage{amsmath,amssymb,amsfonts}
\usepackage{algorithmic}
\usepackage{graphicx}
\usepackage{textcomp}
\usepackage{xcolor}
\usepackage{amsmath}
\usepackage{bm}
\usepackage[font=small, labelfont=bf]{caption}
\usepackage{censor}

\usepackage{url}
\usepackage[hidelinks = True]{hyperref}
\usepackage{multirow}

\def\BibTeX{{\rm B\kern-.05em{\sc i\kern-.025em b}\kern-.08em
    T\kern-.1667em\lower.7ex\hbox{E}\kern-.125emX}}

\graphicspath{{./FinalFigures/}}

\begin{document}

\title{Augmenting Human Balance with \\ Generic Supernumerary Robotic Limbs}

\author{Xuanyun Qiu$^1$, Dorian Verdel$^1$, Hector Cervantes-Culebro$^1$, Alexis Devillard$^1$, and Etienne Burdet$^1$
\thanks{*This work was supported by the H2020 NIMA (FETOPEN 899626), and by the ERC Synergy Natural BionicS (810346).}
\thanks{$^1$All authors are with the Bioengineering Department, Imperial College of Science, Technology and Medicine, W12 0BZ London, United-Kingdom
{\tt\small e.burdet@imperial.ac.uk}}%
}

\maketitle

\begin{abstract}
Supernumerary robotic limbs (SLs) have the potential to transform a wide range of human activities, yet their usability remains limited by key technical challenges, particularly in ensuring safety and achieving versatile control. Here, we address the critical problem of maintaining balance in the human–SLs system, a prerequisite for safe and comfortable augmentation tasks. Unlike previous approaches that developed SLs specifically for stability support, we propose a general framework for preserving balance with SLs designed for generic use. Our hierarchical three-layer architecture consists of: (i) a prediction layer that estimates human trunk and center of mass (CoM) dynamics, (ii) a planning layer that generates optimal CoM trajectories to counteract trunk movements and computes the corresponding SL control inputs, and (iii) a control layer that executes these inputs on the SL hardware. We evaluated the framework with ten participants performing forward and lateral bending tasks. The results show a clear reduction in stance instability, demonstrating the framework’s effectiveness in enhancing balance. This work paves the path towards safe and versatile human-SLs interactions. [This paper has been submitted for publication to IEEE.]
\end{abstract}


\section{Introduction}

\textit{Supernumerary robotic limbs} (SLs), which enable humans to perform tasks with more than two hands, hold the potential to transform how we interact with our environment \cite{Eden2022}--for instance, in industrial assembly or four-hand surgery, where they could reduce reliance on assistants and avoid miscommunication. However, current SLs are limited either to direct control with limbs not involved in the primary task--incurring high cognitive load--or to a few preprogrammed autonomous behaviors \cite{Bonilla2012, LlorensBonilla2013}, which may fail to match user intent, leading to frustration and potential safety risks \cite{Khoramshahi2023}.

To address these limitations, a hierarchical predictive-coding framework for safe and versatile SL operation was proposed in \cite{Verdel_IJRR}. Here, we present an implementation and experimental validation on the basic aspect of operational stability with two wearable SLs. For this purpose, we rely on the unique MUltilimb Virtual Environment (\href{https://www.imperial.ac.uk/human-robotics/dr-octopus-/}{MUVE}), which enables simultaneous measurement of human motion and control of wearable SLs, allowing us to systematically investigate augmented human behavior.

Several pioneering studies have designed SLs dedicated to improving balance. Examples include anchor-type SLs that fix to the environment to ensure stability during aircraft assembly \cite{Parietti2014, Parietti2015, Parietti2016}; gyroscopic wheels \cite{Romtrairat2019, Lemus2020} and robotic tails \cite{Abeywardena2023, Abeywardena2023a} that counteract instability by acting as dynamic counterweights; and supernumerary legs that expand the user’s base of support, effectively transforming bipeds into quadrupeds \cite{Gonzalez2019, Hao2020, Khazoom2020}.

However, the function of these dedicated systems is restricted to stability enhancement. In contrast, our objective is to develop stability strategies for SLs that can support a broad range of functions. In that direction, the study \cite{Moon2024} investigated static postural balancing with multiple SLs on a manikin. Here we introduce a new class of autonomous behaviors that \textit{stabilize the augmented body during movement}. Inspired by human motor control—where unused limbs contribute to postural stabilization \cite{Sainburg2002, Przybyla2013}—we design reflex-like responses for SLs. Such behaviors can enhance safety and comfort in human-SL interaction, preventing injuries to the user and damage to the robots \cite{Verdel_IJRR}.

Implementing such reflexes requires continuous monitoring and prediction of human stance, enabling SLs to plan and execute optimal counteractions efficiently. This is particularly challenging in high-dimensional systems, where real-time computation is needed to proactively assist the user. Importantly, we carried out experiments with ten participants to test and validate the balance behavior.

In the remainder of the paper, we present a general framework for implementing reflex-like stabilization in the presence of uncertainty about the human movements (Section \ref{Sec:Method}). We then describe the experimental design used for evaluation (Section \ref{Sec:Expe}), report results of human–SL interaction (Section \ref{Sec:Results}), and discuss their implications (Section \ref{Sec:Discussion}).

\section{Methodology} \label{Sec:Method}

\subsection{Hierarchical control framework} \label{Subsec:Controller}

\begin{figure}[!ht]
\begin{center}
    \includegraphics[width=0.48\textwidth]{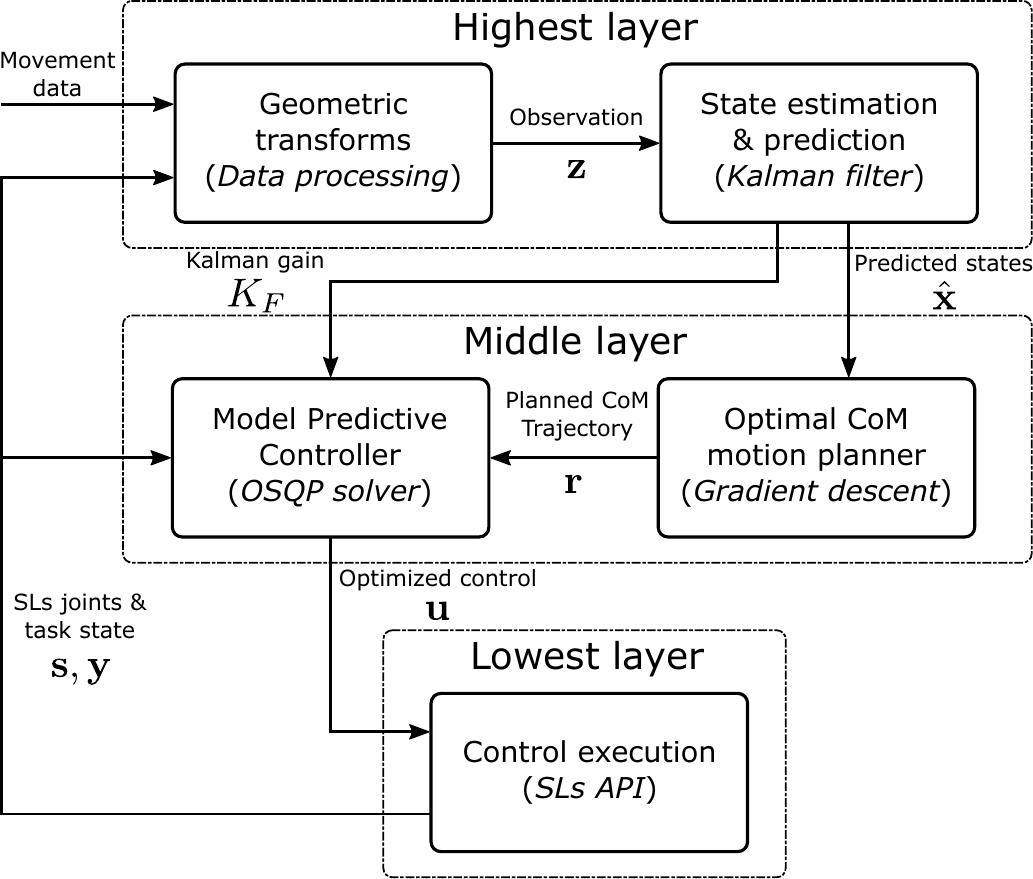}
    \caption{\textbf{Hierarchical control structure for human-SLs balance.} The top layer estimates the human-SLs system state and predicts future states using LQE. The middle layer plans (i) a trajectory that minimizes the shift of the human-SLs center of mass projected in a horizontal plane, and (ii) the future torques required to follow this trajectory. The bottom layer generates motor commands for the SLs.}    
    \label{Fig:Framework}
\end{center}
\end{figure}

We introduce a control framework to enhance human balance when equipped with SLs. Importantly, the specific constraints of SLs--particularly ISO standards for safe collaboration \cite{Verdel_IJRR}--require proactive control methods that compensate for the inherent slowness of the robots by predicting upcoming events. The framework, summarized in Fig.\,\ref{Fig:Framework}, comprises three layers: the bottom layer executes motor commands via the SLs API, while the two upper layers are implemented as described below.

\subsection{State estimation and prediction layer}  \label{Subsec:HighLayer}

\begin{figure*}[htbp]
    \centering
    \includegraphics[width=1\textwidth]{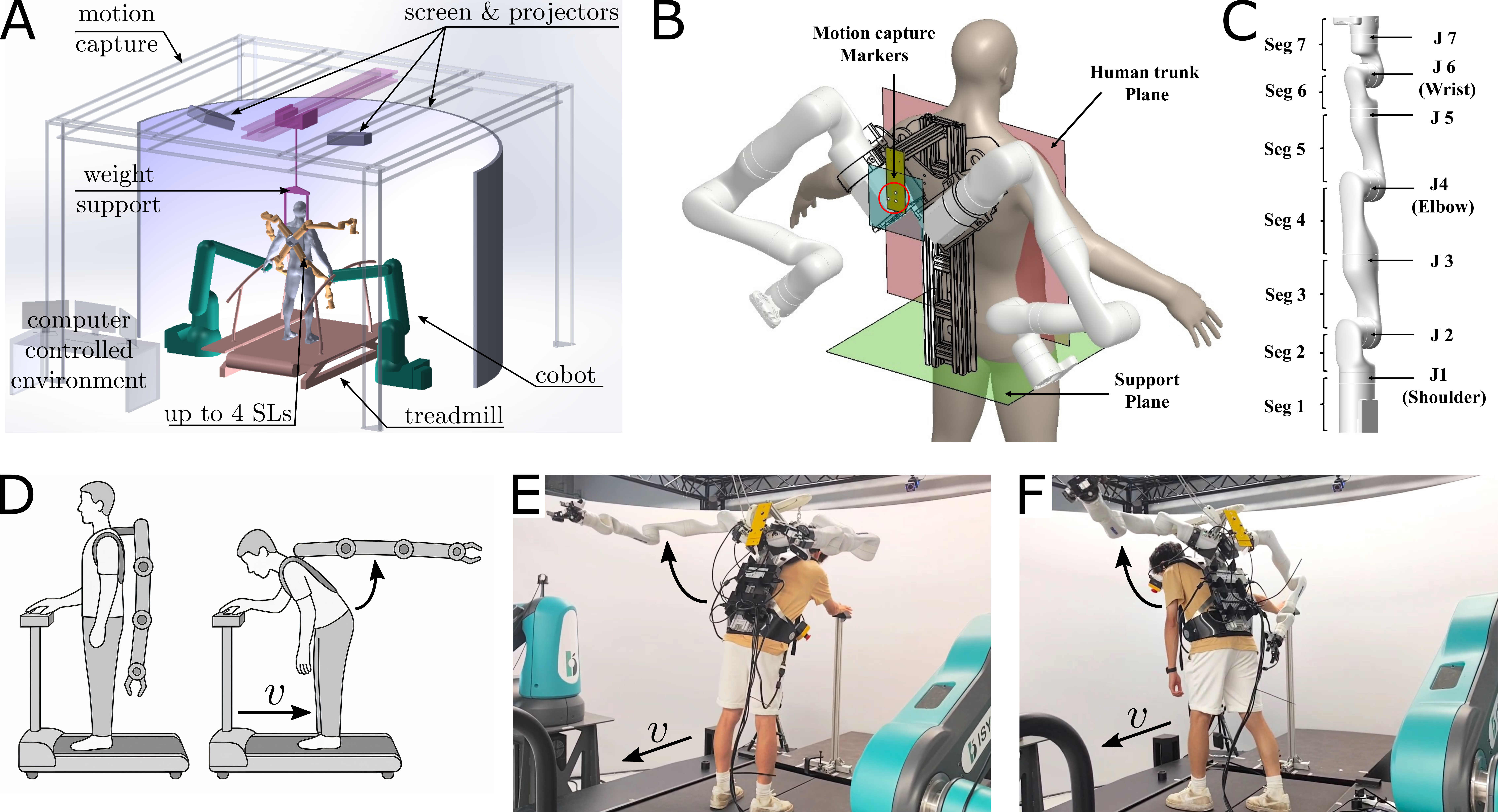}
    \caption{\textbf{Experimental setup.} \textbf{A.} MUVE system integrating motion capture, an orientable split treadmill, up to four SLs and two cobots. \textbf{B.} Definition of the planes used to control the SLs and positions of the motion capture markers used to extract these planes. \textbf{C.} Nomenclature of one SL's segments and joints. \textbf{D.} Protocol used to systematically induce instability in the human-SLs system. \textbf{E,F.} Photos of a representative participant with SLs providing active stabilization during a frontal bow trial (E) and a lateral bow trial (F).}
    \label{Fig:Setup}
\end{figure*}

The top layer estimates the current and future states of the human-SLs system from past observations. For simplicity and computational efficiency, we assume that (i) the human lower limbs remain mostly straight and vertical, as in our validation experiment (Section \ref{Sec:Expe}), (ii) the influence of the arms and head on the overall CoM is negligible, and (iii) the human-SLs connection is rigid. The relevant planes derived from these assumptions are shown in Fig.\,\ref{Fig:Setup}B.

\vspace{0.5em}
\noindent\textbf{Human and SLs CoMs:}
Under these assumptions, trunk orientation is estimated from the backpack orientation using standard geometric transforms. The \textit{center of the human support plane} (SUP) is taken as the average of the left and right hip projections onto the horizontal plane, and the \textit{human center of mass} (HCoM) is the barycenter of the mass-weighted CoMs of the trunk and lower limbs, estimated using anthropometric tables \cite{Winter1990}. Fixed backpack components (e.g., power supply, robot bases) are included in the trunk mass and CoM, and the weight of the SLs last segment is neglected.

We then build a state vector for the coupled human-SLs system comprising (i) the projected position and velocity of the CoM of the human-SLs system $\mathbf{p}\in\mathbb{R}^2$ and $\dot{\mathbf{p}}\in\mathbb{R}^2$, (ii) the projected position and velocity of SUP $\mathbf{p}_{sup}\in\mathbb{R}^2$ and $\dot{\mathbf{p}}_{sup}$, (iii) the projected HCoM position and velocity $\mathbf{p}_{h}\in\mathbb{R}^2$ and $\dot{\mathbf{p}}_{h}$, and (iv) SLs shoulder $\mathbf{p}_j^s\in\mathbb{R}^3$, elbow $\mathbf{p}_j^e\in\mathbb{R}^3$ and wrist $\mathbf{p}_j^w\in\mathbb{R}^3$ joints task space poses with $j\in\{1,2\}$ the SL identifier, and corresponding velocities $\dot{\mathbf{p}}_j^s$, $\dot{\mathbf{p}}_j^e$, and $\dot{\mathbf{p}}_j^w$. The state vector concatenating all these variables is $\mathbf{x}(t)\in\mathbb{R}^{48}$. Note that this state vector is specific to our context and available computational power, but can be extended for higher accuracy or different human-SLs structures.

\vspace{0.5em}
\noindent\textbf{State estimation and prediction:}
Predictive methods are essential to assist balance proactively. One possible approach would be to build a database of human movements and learn predictive models from demonstrations \cite{Calinon2010a, Luo2017, Saveriano2023}. However, collecting such data is challenging, and the effect of human adaptation on prediction quality cannot be evaluated a priori. Instead, we use LQE to estimate the system state $\hat{\mathbf{x}}(t)$ from noisy observations $\mathbf{z}(t)$. This estimate is then used by the middle layer to compute optimal CoM corrections at each timestep.

\subsection{Motion planning layer}

Here, we describe how the predicted states of the human-SLs system $\hat{\mathbf{x}}(t)$, provided by the top layer, are used to plan SLs movements and generate control inputs to improve balance. The second layer consists of a CoM trajectory planner, which provides a reference trajectory to a model predictive controller (MPC) for execution.

\vspace{0.5em}
\noindent\textbf{Optimal CoM trajectory planner:}
We formulate the problem of minimizing the human-SLs instability as an optimal control problem solved at every SLs control timestep (at 1\,kHz in our case). The costs depends on the predicted human SUP and CoM of the human-SLs system.

At each timestep, the problem consists of moving the predicted human-SLs CoM $\hat{\mathbf{p}}(t)\subset\hat{\mathbf{x}}(t)$ to enhance the system's stability through minimizing the cost function
\begin{equation}
V(\mathbf{p}, \dot{\mathbf{p}}) 
= \underbrace{\gamma\|\mathbf{p} - \mathbf{p}_{sup}\|^2}_{\text{CoM shift term}}
+\!\! \underbrace{\zeta\|\dot{\mathbf{p}}\|^2}_{\text{``Effort'' term}}
\label{Eq:Cost_Fun}
\end{equation}
where $\gamma, \zeta > 0$ weigh the human-SLs CoM shift and the CoM velocity (i.e. a measure of effort). Stability of this CoM controller can be shown using Lyapunov theory \cite{Haddad2008}. The minimization is carried out using a gradient descent, yielding an optimal CoM change $\dot{\mathbf{p}}^*$, and the direct formulation in task space allows computation at high rate.

Using the masses and geometric model of the robots, the optimal CoM correction is converted into a reference trajectory $\mathbf{r}(t)$ that minimizes CoM-SUP distance. This reference trajectory is defined in task space through the elbow and wrist of the two SLs as $\mathbf{y}=[\mathbf{p}_1^{e'},\dot{\mathbf{p}}_1^{e'},\mathbf{p}_1^{w'},\dot{\mathbf{p}}_1^{w'},\mathbf{p}_2^{e'},\dot{\mathbf{p}}_2^{e'},\mathbf{p}_2^{w'},\dot{\mathbf{p}}_2^{w'}]'\in\mathbb{R}^{24}$. The shoulders are not included in the planned trajectory since they are fixed relative to the human frame.

\vspace{0.5em}
\noindent\textbf{Model predictive controller:}
The MPC generates SL control inputs to track $\mathbf{r}(t)$. The two SLs are represented in the state vector $\mathbf{s}=[\mathbf{q}_1',\dot{\mathbf{q}}_1',\mathbf{q}_2',\dot{\mathbf{q}}_2']'\in\mathbb{R}^{16}$, which includes their joint positions and velocities, with each SL comprising four active joints. Joints J5--7 (see Fig.\,\ref{Fig:Setup}C) were excluded from the controller because of their negligible influence on the human-SLs CoM. Joint kinematics are written as
\begin{equation}
    \dot{\mathbf{s}} = \mathbf{As}+\mathbf{Bu},
\end{equation}
where $\mathbf{u}\in\mathbb{R}^8$ contains the joint accelerations, and $\mathbf{A}\in\mathbb{R}^{16\times16}$, $\mathbf{B}\in\mathbb{R}^{16\times8}$ encode time derivatives. Differential kinematics map between $\mathbf{s}$ and task state $\mathbf{y}$.

The MPC cost function minimizes task-space tracking error over horizon $\triangle$:
\begin{align}
C(\mathbf{u}) = \int_{t_c}^{t_c+\triangle} \!\!\!\!\!\!\!\!\! &[\mathbf{y}(t) - \mathbf{r}(t)]' \mathbf{Q}_r(t) [\mathbf{y}(t) - \mathbf{r}(t)]     \nonumber
\\
& + \mathbf{u}(t)' \mathbf{R}_r(t) \mathbf{u}(t) + \dot{\mathbf{u}}(t)'\mathbf{W}_r\dot{\mathbf{u}}(t)\, dt
\end{align}
where $t_c$ is the current time, $\mathbf{Q}_r(t) \in \mathbb{R}^{24 \times 24}$ penalizes tracking errors with respect to the reference trajectory  $\mathbf{r}(t)$ of the SLs elbow and wrist in task space. $\mathbf{R}_r(t) \in \mathbb{R}^{8 \times 8}$ penalizes acceleration and $\mathbf{W}_r \in \mathbb{R}^{8 \times 8}$ jerk, thereby ensuring smooth planned motion.

To enhance the robustness of our controller to state estimation uncertainties, the weighting matrices are adapted using the Kalman gain Frobenius norm $K_F\triangleq\|\mathbf{K}(t)\|_F$,
\begin{equation}
\mathbf{Q}_r(t) = \, \mathbf{Q}_r^0 \! \left(\! 1 \!-\! \frac{K_F}{K_{\text{0}}} \! \right), \quad \mathbf{R}_r(t) = \,\mathbf{R}_r^0 \! \left(\! 1 \!+\! \frac{K_F}{K_{\text{0}}} \! \right)
\end{equation}
with $K_{\text{0}}\triangleq4$ chosen empirically to trade-off responsiveness and stability. This strategy reduces the tracking gains while simultaneously increasing the control input costs when the uncertainty regarding the estimated state increases, thereby improving safety in presence of uncertainty.

The MPC is formulated as a quadratic program (QP) with bounds on joint positions, velocities, and accelerations. We solve it using OSQP, which exploits problem sparsity for real-time convergence. Warm-starting with the previous solution further reduces computation time.

\section{Material and Experiment} \label{Sec:Expe}
We designed an experiment to evaluate whether the SLs controller presented in Section\,\ref{Subsec:Controller} could efficiently reduce a user's instability.

The experimental protocol was approved by the Imperial College London ethics committee (SETREC, 7111986). 10 healthy right-handed males (age 26.6$\pm$6.6\,years, height 174$\pm$ 9\,cm, weight 73.7$\pm$ 12.8\,kg) participated in the experiment. All participants provided written informed consent prior to the experiment.

\textit{The MUltilimb Virtual Environment} (MUVE, Fig.\,\ref{Fig:Setup}A) enables synchronous interaction with up to four SLs mounted on the user’s back. In this experiment, two SLs were used (7-DOF, Kinova Gen3, Boisbriand, Canada), resulting in a backpack weight of approximately 30 kg. The platform also includes two cobots (not used here) positioned on either side of a treadmill equipped with force plates (GRAIL platform, Motek, Houten, Netherlands). To ensure participant safety, the backpack was suspended from the ceiling by cables that prevented falls without affecting the experiment, and emergency stop buttons cutting the SLs power were available for the participant. MUVE is equipped with a motion capture system (VICON, Oxford, UK) recording at 100\,Hz, used to extract the SUP and backpack planes (Fig.,\ref{Fig:Setup}B).

\textit{Ground reaction forces} (GRF) and the center of pressure (CoP) were obtained from the treadmill force plates, also sampled at 100\,Hz. Motion capture and force data were low-pass filtered using a first-order Butterworth filter with a 2 Hz cut-off frequency.

\textit{Through the experiment}, participants were systematically placed in unstable postures by making them bend forward (\textit{frontal bow}) or rightward (\textit{lateral bow}), as illustrated in Fig.\,\ref{Fig:Setup}D-F. In frontal bow trials, participants stood upright at the front of the treadmill, pressing a button with their right index finger. The treadmill then moved backward at a constant velocity of 0.04\,m/s for 7.5\,s, or until the participant lifted a foot or could no longer press the button. In lateral bow trials, participants began rotated 90$^\circ$ leftwards, pressing a button located to their right. Their feet were aligned in a straight line, creating a narrow support polygon and an inherently unstable posture from trial onset. Each participant performed five trials per condition, repeated three times, in the following order: (i) without SLs (\textit{HOnly}), (ii) with inactive SLs (\textit{NoComp}), and (iii) with SLs controlled by the framework in Fig.,\ref{Fig:Framework} (\textit{Comp}). In total, each participant completed 30 trials, 20 of which were performed with the SLs.

\textit{In terms of statistics}, as a Shapiro-Wilk test indicated that our data was not normally distributed, we used non-parametric statistical tests to compare balance between different conditions. Specifically, main effects were assessed using Friedman tests, with Kendall's W as the effect size. When significant, post-hoc comparisons were performed with Wilcoxon signed-rank tests, and the Cohen's D was reported as the effect size. The significance of all tests was set at $p<0.05$.

\section{Results}   \label{Sec:Results}

We evaluated the participants' balance during the trials as the distance between center of mass (CoM) of the human\&SLs system and the human support center (SUP), and using ground reaction forces (GRF). Balance is improved when the CoM-SUP distance is small, then the user would need little horizontal GRF. 
\begin{figure*}[htbp]
    \centering
    \includegraphics[width=1\textwidth]{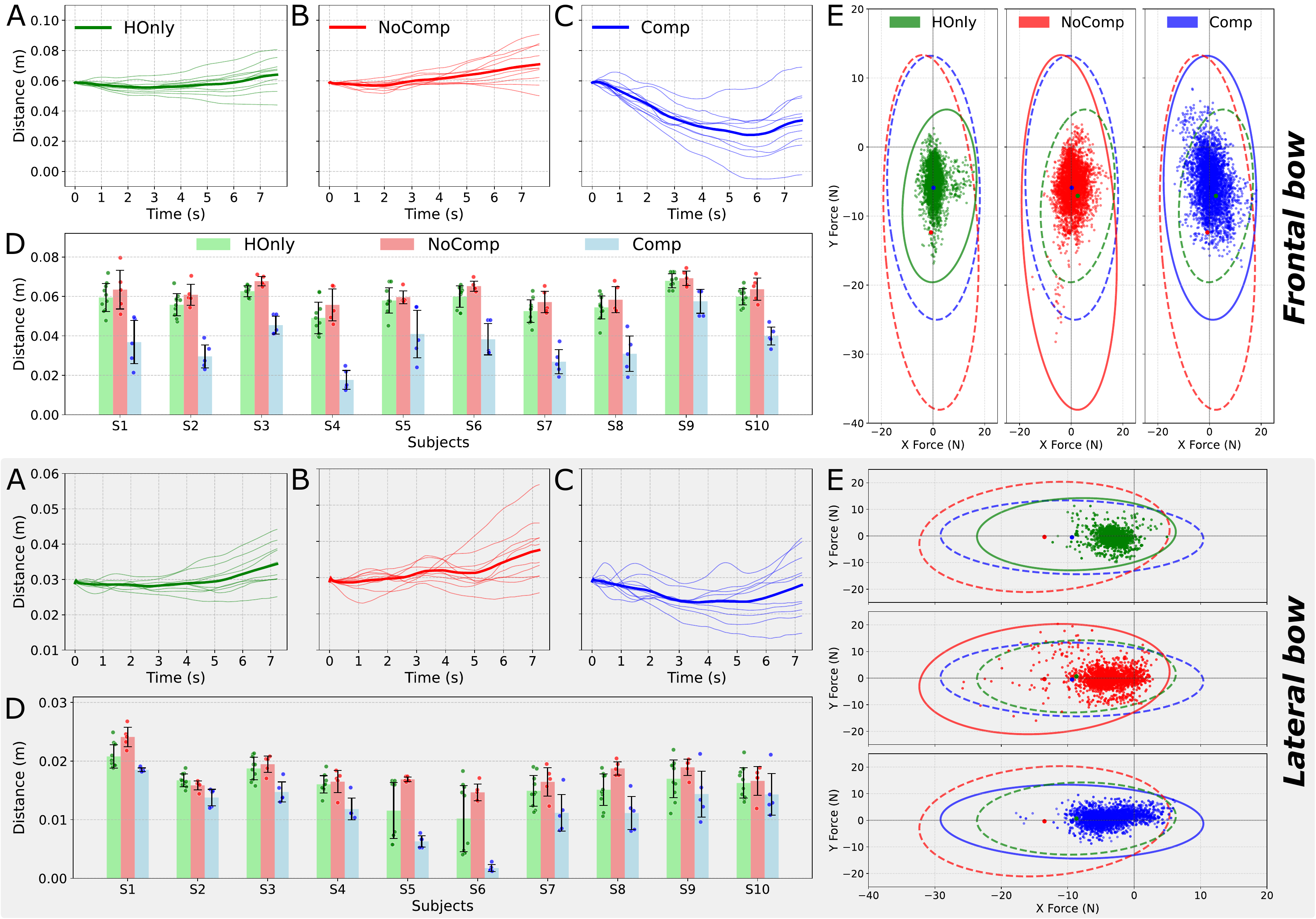}
    \caption{\textbf{Results in \textit{frontal bow} trials (top) and \textit{lateral bow} trials (bottom).} \textbf{A-C.} Evolution of the CoM-SUP distance for each individual (thin lines) and the population average (thick line), in the HOnly, NoComp and Comp conditions respectively. \textbf{D.} Average CoM-SUP distance across the three conditions for all ten participants. \textbf{E.} Ground reaction force (GRF) in the three conditions. Ellipses indicate the maximum GRF of the population average, and the dots show the force distribution for a representative participant.}
    \label{Fig:CoMSUP_Both}
\end{figure*}

First, the results of the \textit{frontal bow} trials are summarized in Fig.\,\ref{Fig:CoMSUP_Both}. Fig.\,\ref{Fig:CoMSUP_Both}A-C exhibits clear differences in the time evolution of the CoM-SUP distance for the different conditions. In HOnly, the CoM-SUP distance tends to increase as the treadmill moves backward and the bending motion increases. In NoComp, the trend is similar to HOnly but with a more pronounced increase, suggesting a detrimental impact of the passive SLs. However, in Comp, the CoM-SUP distance clearly decreases throughout the movement, with a minimum usually reached within 6.5\,s after the beginning of the trial. This suggests that our controller increases balance up to a certain bending level, after which the mass of the SLs becomes insufficient to counteract the weight of the human.

These results are confirmed by a significant main effect of the condition on the average CoM-SUP distance (see Fig.\,\ref{Fig:CoMSUP_Both}D) during the \textit{frontal bow} trials ($\chi^2_2=20$, $p<10^{-4}$, Kendall's $W=1$). Specifically, post-hoc comparisons show a better human-SLs balance in Comp than in both HOnly and Uncomp (in both cases: $p<10^{-2}$, Cohen's $D>2.49$), and a detrimental effect of the passive SLs compared to HOnly ($p<10^{-2}$, Cohen's $D=0.84$).

Finally, the horizontal components of the GRF reported in Fig.\,\ref{Fig:CoMSUP_Both}E, and fitted maximum force ellipses, show that our control framework tends to decrease the asymmetry of the force distribution around 0\,N when compared to HOnly and NoComp, which is consistent with a more balanced human-SLs system and lower maximum instability.

The \textit{lateral bow} trials in Fig.\,\ref{Fig:CoMSUP_Both} exhibits similar results. Fig.\,\ref{Fig:CoMSUP_Both}A-C show that the HOnly trials lead to a gradual decrease in balance, which is more pronounced when wearing SLs in NoComp, while our controller seemingly leads to improved balance in Comp. These trends are confirmed by a significant main effect of the condition during \textit{lateral bow} trials ($\chi^2_2=18.2$, $p<10^{-4}$, Kendall's $W=0.91$). Specifically, as for the \textit{frontal bow} trials, post-hoc comparisons showed a better human-SLs balance in Comp than in both HOnly and Uncomp (in both cases: $p<10^{-2}$, Cohen's $D>0.98$), and a detrimental effect of the passive SLs compared to HOnly ($p=0.016$, Cohen's $D=0.72$). Finally, the horizontal components of the GRF reported in Fig.\,\ref{Fig:CoMSUP_Both}E, and fitted maximum force ellipses, show that our control framework tends to decrease the asymmetry of the force distribution around 0\,N when compared to HOnly and NoComp, which is similar to \textit{frontal bow} trials, and consistent with a more balanced human-SLs system and lower maximum instability.

\begin{figure}[!ht]
\begin{center}
    \includegraphics[width=0.5\textwidth]{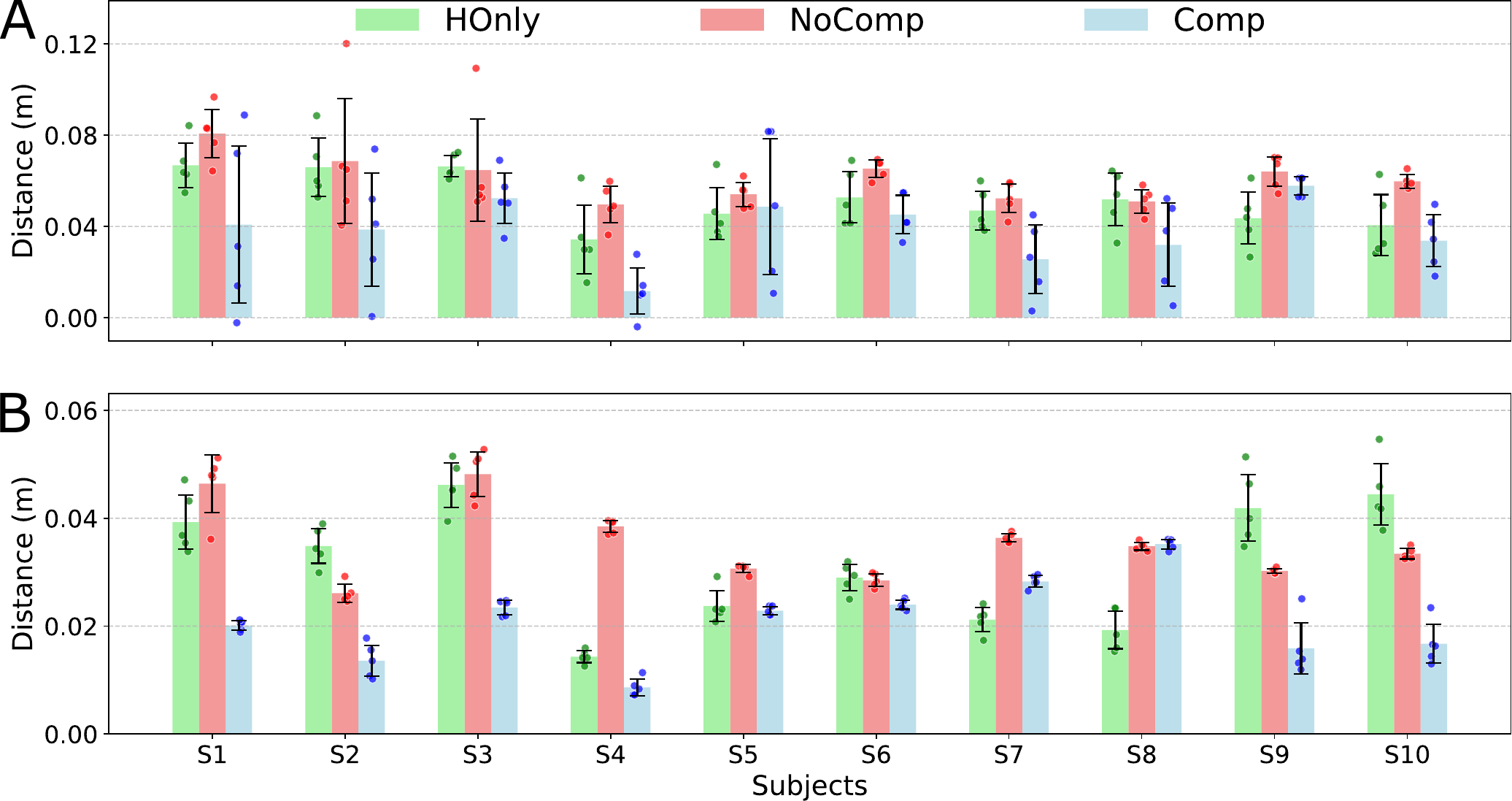}
    \caption{\textbf{CoP-SUP distance.}  Average distance across each trial for every participant in \textbf{A} frontal bow trials and \textbf{B} lateral bow trials.}
    \label{Fig:CoPSUP_Both}
\end{center}
\end{figure}

Importantly, since computing the CoM of the human-SLs system required to use anthropometric models and manufacturer's data that can be inaccurate, we also evaluate the effect of our controller on the distance between the CoP measured by the force plates and SUP. The averaged CoP-SUP distance obtained for each trial and participant are reported in Fig.\,\ref{Fig:CoPSUP_Both}. For both \textit{frontal bow} ($\chi^2_2=12.8$, $p<10^{-2}$, Kendall's $W=0.64$) and \textit{lateral bow} ($\chi^2_2=7.8$, $p=0.02$, Kendall's $W=0.39$) trials, we find a significant main effect of the condition on the CoP-SUP distance. In \textit{frontal bow} trials, Comp leads to a lower CoP-SUP distance when compared to both HOnly and NoComp (in both cases: $p<0.03$, Cohen's $D>1$), while NoComp led to a significantly worse stability than HOnly ($p=0.013$, Cohen's $D=0.9$). Differences are less pronounced in the \textit{lateral bow} trials, as Comp led to significantly better stability when compared to NoComp ($p<10^{-2}$, Cohen's $D=1.9$) but not when compared to HOnly despite a large effect size and visible trend ($p=0.06$, Cohen's $D=1$).

\section{Discussion}    \label{Sec:Discussion}

This paper introduced and implemented a hierarchical control framework, part of recent conceptual developments \cite{Verdel_IJRR}, to increase balance while using worn supernumerary robotic limbs (SLs). The framework relies on predicting human trunk movements with LQE, and optimally compensating for the resulting human-SLs CoM displacements. Unlike static stability compensation approaches such as \cite{Moon2024}, our framework adapts in real time to arbitrary trunk movements, since both the desired CoM trajectory and associated control inputs are re-optimized at high frequency in an MPC fashion. Moreover, because our method focuses on compensating CoM motion of the overall human–SLs system, it could readily be employed even when one SL is engaged in another task. Balance augmentation with generic SLs as described in the present paper--adapted to assist users during volitional posture selection and maintenance while performing a task with SLs--are complementary to highly reactive techniques based on gyroscopic wheels \cite{Romtrairat2019, Lemus2020}--adapted to prevent falls and react to external disturbances but are not suitable to provide sustained assistance during a task. We validated this approach in an experiment with ten participants using generic wearable SLs that can serve multiple purposes. The results demonstrated the benefits of augmented balance, even when compared to a human without SLs. These findings support the concept of \textit{augmented human balance}.

In this work, we simplified both human modeling and control to enable fast computation suitable for our validation experiment. Other applications may require including the head and arms in the human model or exploiting all SL joints. While such extensions would not alter the control principle, they would significantly increase the computational burden, thus motivating more efficient optimization methods.

A further limitation of our system lies in its weight, which makes prolonged use impractical. The current SLs backpack weighs around 30\,kg, with a usable payload of 8\,kg when using both arms together. More than half of this weight comes from the robot arms (8\,kg each) and the embedded electrical converters (2\,kg each), not including the battery. Lighter supernumerary arms, such as the Jizai arms \cite{Yamamura2023}, have been developed, but they are far less powerful and thus not suitable for practical tasks. Furthermore, overly light arms may compromise balance compensation. While the arm mass cannot be reduced excessively, a lighter backpack could be designed. Our current backpack system was built to support up to four SLs and modular configurations, which added considerable weight.

Future research on SL control should focus on integrating balance augmentation methods such as those presented here, which constitute basic autonomous behaviors, with advanced interactive control methods that blend voluntary control and autonomy. Promising directions include leveraging Lipschitz analysis \cite{Song2021, Song2023} and multi-agent–based control \cite{Verdel_IJRR}. These approaches could help achieve the objectives of robotic augmentation in industrial and healthcare applications identified over the past decade \cite{Yang2021a, Eden2022}.

\bibliographystyle{ieeetr}
\bibliography{Biblio_these}

\end{document}